\documentclass[conference]{IEEEtran}
\IEEEoverridecommandlockouts
\usepackage{cite}
\usepackage{amsmath,amssymb,amsfonts}
\usepackage{graphicx}
\usepackage{textcomp}
\usepackage{xcolor}
\usepackage{tikz}
\usetikzlibrary{arrows.meta,positioning}
\usepackage{booktabs}
\usepackage{array}

\def\BibTeX{{\rm B\kern-.05em{\sc i\kern-.025em b}\kern-.08em
    T\kern-.1667em\lower.7ex\hbox{E}\kern-.125emX}}

\begin{document}

\title{AI-accelerated End-to-End Framework for \\Rapid
Professional Upskilling}

% NOTE: second author's name carries Vietnamese diacritics (Nguyen with
% breve/tilde). Rendered here in ASCII for pdfLaTeX safety; to typeset
% "Nguy\~{\^e}n" exactly, add \usepackage[T5]{fontenc} and test on Overleaf.
\author{
\IEEEauthorblockN{Tam Nguyen}
\IEEEauthorblockA{\textit{Crew Scaler / US Federal Government}\\
Washington, DC, USA\\
T@crewscaler.org}
\and
\IEEEauthorblockN{Hung Nguyen}
\IEEEauthorblockA{\textit{Crew Scaler}\\
Boston, MA, USA\\
hungtrung.nguyen@crewscaler.org}

\and
\IEEEauthorblockN{Robert Ogburn}
\IEEEauthorblockA{\textit{Crew Scaler / US Federal Government}\\
Palm Coast, FL, USA\\
R@crewscaler.org}

\thanks{Corresponding author: Tam Nguyen (t@crewscaler.org).}
}

\maketitle

\begin{abstract}
By 2030, 59 of every 100 workers will need reskilling or upskilling, yet the average time to close an enterprise skills gap grew from roughly 3 days in 2014 to 36 days in 2018. Most current frameworks accelerate single stages of upskilling programs and generally lack industry validation. We present an end-to-end framework that applies AI acceleration across five stages of knowledge acquisition, content development, content review and verification, teaching, and assessment development; with a strong focus on both production and learning efficiency. Three strong external signals validates the framework: the US National Association of State Boards of Accountancy reviewed and approved an upskilling program built on the framework for continuing-professional-education credits; 3 learners followed the program and passed the NVIDIA Certified Professional in Agentic AI exam in a significantly short amount of time, with 14 more in progress; the program's knowledge base supports complex downstream analysis such as the production of a robust 1,267 risk item dataset for managing multi-agent AI system risks. 
\end{abstract}

%can you see this now?%%%%%

\section{Introduction}
\IEEEPARstart{F}{ifty-nine} of every 100 workers will need reskilling or upskilling by 2030, and 11 of them are unlikely to receive it. In other words, roughly 120 million workers are at the medium-term risk of redundancy~\cite{wef2025}. The same forecast expects 39\% of core skills to change or become outdated by 2030 as 170 million jobs are created and 92 million displaced~\cite{wef2025}. The skills themselves decay. Skill relevance averages about five years, while that of technical skills is about two and a half~\cite{daniel2020halflife}. Organizational response is moving the other way. The average time to close an enterprise skills gap through classroom or online training rose from roughly 3 days in 2014 to 36 days in 2018~\cite{laprade2019}. The result is a mismatch paradox as layoffs coexist with unfilled vacancies because required competencies change faster than workers adapt. The binding constraint is no longer whether to upskill but how fast rigorous upskilling programs can be produced and executed.

Generative AI sharpens the problem because its benefits are uneven. An assistant deployed to 5,179 customer-support agents raised issues resolved per hour by 15\% on average but by 30\% for less skilled and less experienced workers~\cite{brynjolfsson2025genai}; among 453 professionals, ChatGPT helped initially weaker writers most~\cite{noy2023science}; and a field experiment with 758 consultants found gains concentrated among lower-scoring participants~\cite{dellacqua2023jagged}. We interpret this pattern through a two-regime framing of the cited results. When work reduces to fixed prompts with easy-to-judge outputs, AI acts as a \emph{leveler} and novices match experts. When work demands flexible prompting and carries a high evaluation burden, AI acts as a \emph{multiplier} and expertise compounds. Therefore, how a workforce is upskilled shapes which regime it occupies.

The stakes are institutional as much as individual. In a survey of 116 executives at large organizations, roughly one in four lacked a clear view of how automation will affect skill requirements, and nearly one-third doubted their HR infrastructure could execute a skills strategy~\cite{mckinseyretraining}. Institutions that cannot produce rigorous, judgment-building upskilling programs at the pace of their AI exposure will absorb that exposure unprepared.

We present a framework that closes this production-capability gap by applying AI acceleration to every stage of upskilling-program rather than to just one or few single stage. As Section~II details, existing frameworks accelerate few single stages and generally lack external industry validation. We are explicit about the nature of our evidence: we report design, artifacts, and external validation signals, not controlled-comparison evidence. The paper contributes (C1) an end-to-end AI-accelerated production pipeline for certification-grade programs (aligned to a reputable vendor's certification blueprint and sufficient to prepare candidates for its exam) spanning knowledge acquisition, content development, content review and verification, AI-tutor coaching, and assessment development; (C2) a dual-efficiency design pairing AI-accelerated production with learning-efficient outputs; (C3) specific external validation signals including: approval by NASBA (the National Association of State Boards of Accountancy) for continuing-professional-education (CPE) credits; certification passes (3 learners, 14 in progress); downstream capability validated by institutional subject-matter-expert (SME) reviews; and (C4) documented stage-level methods enabling replication. Section~II reviews related work and gaps, Section~III describes the framework's five stages, Section~IV reports the validation signals, Section~V discusses implications, and Section~VI concludes.

\section{Background and Related Work}

Adult professionals do not learn the way children do. Knowles
characterizes adults as self-directed, experience-anchored, problem-centered
learners who require instruction built around authentic professional
tasks~\cite{knowles2015adult}. Yet AI educational technologies are designed
and evaluated predominantly for K-12 contexts and remain ``poorly
aligned'' with adult learners' needs~\cite{reddig2026guidelines}, and the
LLM planning tools targeting self-directed learners face documented
transparency and hallucination problems~\cite{chun2025planglow}. 

Our framework is built different. It begins with our fundamental understanding and our definition of \emph{rapid upskilling} - minimizing time-to-competency for professionals acquiring a frontier technical
topic - where competency is demonstrated against an external standard rather than self-reported.

\subsection{Frameworks for Rapid Upskilling}
Existing frameworks accelerate one of four loci: content delivery,
content production, the learning process, or the program. Hence, their
evidence is correspondingly partial.

Delivery-side approaches compress instruction into short units, but the
gains fade. In a controlled comparison, ChatGPT and Google users
outperformed controls on immediate lower-order tasks, yet retention scores
converged to an e-textbook group's~\cite{akgun2025retention}. Compressed
delivery therefore accelerates exposure, not competency.

Production-side frameworks automate the creation of materials.
Instructional Agents generates end-to-end course materials with a
multi-agent LLM pipeline~\cite{yao2025instructional}; a related system
casts LLM agents as learning designers in a five-stage
pipeline~\cite{wang2025masdesigners}; and ARCHED, a human-centered pipeline
guided by Bloom's taxonomy, produced learning objectives statistically
indistinguishable from expert-written ones~\cite{li2025arched}. However, none verifies content at the claim level, faces learners, or validates against an external certification standard.

Learning-side frameworks accelerate the learner rather than the materials.
RAG-PRISM delivers adaptive retrieval-augmented tutoring for rapid
workforce development~\cite{raul2025ragprism}. Tutor CoPilot, a randomized
controlled trial of a human-AI system in live tutoring, raised topic
mastery by 4 percentage points, and by 9 for students of lower-rated
tutors~\cite{wang2024copilot}. Deployed systems corroborate the approach:
a classroom dialogue system kept 71\% of conversations fully
on-track~\cite{liu2026dialogue}, and LLM-generated feedback in introductory
programming raised the odds of eventual
correctness~\cite{heickal2026feedback}. Learner-side acceleration works but
presupposes existing materials.

Program-level efforts span the lifecycle with human labor: the AI
Technicians program - a rapid occupational AI training with the US Army -
succeeded under conditions its authors call ``a cost that other
organizations may not be able to bear''~\cite{savelka2025aitech}.
A systematic review of LLMs in programming education confirms the field
remains a collection of point solutions~\cite{zhu2025llmreview}.

\subsection{Gap Analysis}
Our survey exposes four gaps.

First, the landscape is fragmented. For example, production pipelines stop at
materials~\cite{yao2025instructional, wang2025masdesigners, li2025arched}.
No surveyed framework covers end-to-end results, from knowledge acquisition through successful industry assessment.
Section~III presents an integrated pipeline covering all five stages.

Second, verification is missing where it is most needed. Averaged across
models, LLMs hallucinated on 60.33\% of post-knowledge-cutoff
questions~\cite{alessa2025bias}. Detection is feasible: FAVA pairs a hallucination taxonomy with a
retrieval-augmented detector~\cite{mishra2024fava}, and VeriScore verifies
extracted claims with a fine-tuned
verifier~\cite{song2024veriscore}. Yet no surveyed education pipeline
incorporates such a layer. Section~III-C answers with a dedicated
verification stage.

Third, default LLM pedagogy is shallow. In a 223-domain tutoring testbed,
LLMs produced correct next-step actions only 52--70\% of the
time~\cite{weitekamp2025tutorgym}; the deployed successes above engineered
pedagogy deliberately~\cite{wang2024copilot, heickal2026feedback}. Section~III-D
answers with explicitly designed coaching protocols.

Fourth, outcomes are rarely measured against anything external, let alone industry certification exams on frontier topics. A few reasons include: GenAI evaluation practice lacks scientific rigor~\cite{wallach2025measurement},
zero-shot prompting underperformed traditional NLP in curricular
analytics~\cite{xu2025curricular}, and circular LLM-as-judge validation is
a documented validity threat~\cite{thomas2026groundtruth}. A framework that
generates its own success measure proves nothing.
Sections~III-E and IV answer with blueprint-aligned assessment and
validation against an external certification body.

\section{The Crew Scaler Framework}

\begin{figure}[t]
\centering
\begin{tikzpicture}[node distance=2mm,
  every node/.style={font=\scriptsize},
  stage/.style={draw, rounded corners, align=center, minimum height=7mm,
    inner sep=1.5pt, text width=13.5mm}]
\node[stage] (s1) {Knowledge\\Acquisition};
\node[stage, right=of s1] (s2) {Content\\Development};
\node[stage, right=of s2] (s3) {Content\\Review \&\\Verification};
\node[stage, right=of s3] (s4) {AI-Tutor\\Coaching};
\node[stage, right=of s4] (s5) {Assessment\\Development};
\draw[-Stealth] (s1) -- (s2);
\draw[-Stealth] (s2) -- (s3);
\draw[-Stealth] (s3) -- (s4);
\draw[-Stealth] (s4) -- (s5);
\node[draw, align=center, inner sep=2pt, text width=72mm,
  above=3.5mm of s3] (hum)
  {Human judgment: blueprints, SME review, misconceptions, item ratings};
\draw[dashed, -Stealth] (hum.south -| s1) -- (s1.north);
\draw[dashed, -Stealth] (hum.south) -- (s3.north);
\draw[dashed, -Stealth] (hum.south -| s4) -- (s4.north);
\draw[dashed, -Stealth] (hum.south -| s5) -- (s5.north);
\node[draw, align=center, inner sep=2pt, text width=72mm,
  below=3.5mm of s3] (ext)
  {External checks: NCP-AAI exam; risk-taxonomy derivation; federal SME
   review; NASBA CPE review};
\draw[dashed, -Stealth] (s1.south) -- (ext.north -| s1);
\draw[dashed, -Stealth] (s3.south) -- (ext.north);
\draw[dashed, -Stealth] (s5.south) -- (ext.north -| s5);
\end{tikzpicture}
\caption{The five-stage pipeline. AI accelerates production within each
stage; humans retain high-judgment roles (top); outputs face external
checks (bottom); all learner-facing stages draw on the stage-one knowledge
substrate. NCP-AAI: NVIDIA Certified Professional in Agentic AI.}
\label{fig:pipeline}
\end{figure}
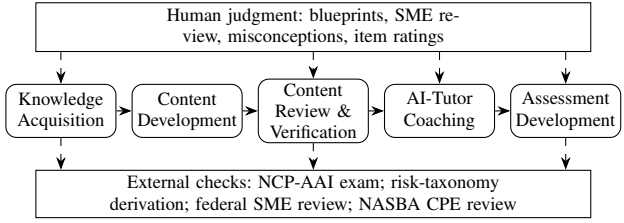

The framework organizes rapid upskilling as a five-stage pipeline: knowledge
acquisition, content development, content review and verification, AI-tutor
teaching, and assessment development (Fig.~\ref{fig:pipeline}). The pipeline is rapid for two reinforcing
reasons. \emph{Production efficiency} means AI compresses the time required
to produce instructional materials at every stage. \emph{Learning
efficiency} means the outputs themselves
are structured to reduce learner time-to-competency through prerequisite
ordering, spaced review, misconception-keyed distractors, and adaptive
tutoring. Humans retain the roles where judgment carries the most
weight such as blueprint design, subject-matter-expert (SME) review, misconception
authoring, and item rating. In the mean time, AI absorbs the volume work those
judgments govern, keeping human expertise in the multiplier regime described
in Section~I. Each stage therefore pairs an AI-acceleration mechanism with a
learning-efficiency mechanism and a quality-control check
(Table~\ref{tab:stages}). Finally, the pipeline's outputs face external validation
the authors do not control (Fig.~\ref{fig:pipeline}, bottom).

\begin{table}[t]
\caption{Stage-level summary: each stage pairs an AI-acceleration mechanism with a learning-efficiency mechanism and a quality control}
\label{tab:stages}
\centering
\small
\begin{tabular}{p{1.55cm}p{2.1cm}p{2.1cm}p{1.6cm}}
\toprule
Stage & AI acceleration & Learning-efficiency mechanism & Quality control \\
\midrule
Knowledge acquisition & LLM-assisted domain exploration, extraction &
Prerequisite-ordered 4-level hierarchy & Blueprint coverage checks \\
Content development & AI-drafted chapters; condensation passes &
One-new-element pacing; 70/20/10 review & Fixed templates; six-pass
revision \\
Content review \& verification & Automated hallucination, faithfulness checks &
Defects caught before learners study & SME audit; immutable audit trail \\
AI-tutor coaching & Scalable one-to-one protocolized tutoring &
Intent- and affect-adaptive protocols & Integrity guardrails; grounded
RAG \\
Assessment development & AI-generated items and distractors &
Misconception-targeted diagnostic distractors & Blueprint tagging;
difficulty distribution \\
\bottomrule
\end{tabular}
\end{table}

\subsection{Knowledge Acquisition}

We surveyed 27 literature-survey for general knowledge domain and 26 threat-modeling methods targeting a specialized knowledge domain before
designing this stage. None was simultaneously scalable and
evidence-grounded~\cite{nguyen2026mas}. That gap motivated our
\emph{Knowledge Domain Exploring Guide} (KDEG), a four-step procedure:
domain framing, practice-grounded job-task analysis, blueprinting into
weighted content domains with a table of specifications, and evidence
mapping that links each objective to authoritative references. AI
accelerates the two labor-dominated steps of domain exploration and
knowledge-item extraction against the blueprint, while humans retain framing, weighting, and evidence admission.
Quality control is structural rather than post hoc. Specifically, extracted content
occupies a four-level dependency hierarchy - foundational concepts, primary
building blocks, integrated concepts, and applied or advanced
material - linked by strict dependency chains. Because material arrives in
prerequisite order, the hierarchy bounds the context a learner needs at each
step to concepts already mastered, and coverage gaps surface against the
blueprint rather than through inspection. Applied to an emerging and uniquely complex knowledge domain of multi-agent-systems, the stage produced the approximately 3,000-page knowledge base.

\subsection{Content Development}

Knowledge base chapters originate as AI-generated drafts and pass through three disciplined
transformations. Structural drafting aligns each draft to the human currated knowledge domain
blueprint through a fixed skeleton (Overview, Learning Objectives, Content,
Key Concepts, Assessment) carrying 3 to 8 SMART learning objectives per
chapter, hands-on labs, and exam-style practice items. A condensation pass
applies named compression techniques such as minimalist documentation,
information mapping, worked-example fading rather than ad hoc cutting. AI
performs the bulk drafting while the fixed templates hold the pedagogical
shape constant, so human effort concentrates on judging fit against the
blueprint. A beginner-readability pass then enforces rules diagnosed from
defects in actual drafts. For example, the ``One New Element'' rule permits exactly one
new complexity per section, and cumulative review distributes roughly 70\%
current-chapter, 20\% prior-chapter, and 10\% foundational questions,
operationalizing spaced retrieval practice~\cite{roediger2006retrieval}. A six-pass revision checklist closes
each chapter.

\subsection{Content Review and Verification}

LLMs hallucinate on a majority of questions past their knowledge
cutoff~\cite{alessa2025bias}. AI-drafted content therefore requires its own
verification machinery. The framework specifies three layers: automated
detection, expert human review, and immutable audit documentation. Layer~1
adapts the RAGAS framework into a four-part accuracy
standard of faithfulness, context precision, answer relevancy, and
harmfulness. To further improve issue detection, the approach uses a four-type hallucination taxonomy: factual, reasoning, contextual, and true fabrications. This supports research showing that both fine-grained taxonomies and claim-level 'extract-then-verify' pipelines enhance accuracy~\cite{mishra2024fava,song2024veriscore}. We executed Layer~1 as a scripted pass producing six reports scored against explicit numeric targets,
and the results were mixed: platform accuracy scored 95\% on a 20-example
sample and cross-chapter integrity found 0 invalid references among 268
checked; conversely, only 5 of 10 chapters met the 70\%
pedagogical-progression bar, and assessment materials stood at 63\% of the
assessment-bank target at that time. We report this mixed picture
deliberately: criterion-referenced checks that can fail (and did) are
evidence that the instrument measures rather than
ratifies~\cite{wallach2025measurement}. Both gaps entered a tracked
remediation backlog; the assessment gap was subsequently closed (the
completed 530-question bank postdates that pass), while the
pedagogical-progression remediation remains tracked. Layer~2 SME sign-off is
fully specified but not yet fully evidenced in execution records. %% i will provide more evidence on this %%

\subsection{AI-Tutor Coaching}

Step-based tutoring approaches human-tutor effect sizes
($d=0.76$)~\cite{vanlehn2011tutoring}, yet in a 223-domain testbed current
LLMs labeled incorrect learner actions no better than
chance~\cite{weitekamp2025tutorgym}, and recent work narrows such gaps by
training tutors against student-outcome
objectives~\cite{scarlatos2025training}. We therefore specified the tutor's
pedagogy in advance as a library of 16 named protocols rather than a single
general-purpose prompt, grounded in the retrieval-practice,
productive-failure, and affect
literatures~\cite{roediger2006retrieval,kapur2014productive,dmello2012affect}. These protocols include:
direct explanation, Socratic questioning, worked examples, hint escalation,
spaced retrieval, productive failure, affective support that prioritizes
boredom over frustration, and an academic-integrity guardrail. A selection
layer picks the active protocol each turn by fixed priority: integrity risk
first, then affective signals, then group context, then learner intent. Six
cross-cutting principles constrain every protocol: knowledge grounding,
learner modeling (formalizable through knowledge
tracing~\cite{corbett1995kt}), verify-before-trusting, ask-don't-tell,
calibrated intervention~\cite{borchers2026engagement}, and step-level
feedback. Error diagnosis serves learning efficiency. Specifically, the tutor classifies
a wrong answer as conceptual, procedural, factual, or careless, then checks
it against a per-chapter misconception catalog, so feedback names the
specific misconception rather than a generic error.

\subsection{Assessment Development}

In a field study across 91 classes, AI-generated questions performed
comparably to expert-created standardized-exam items under
item-response-theory analysis~\cite{isley2025exams}. Our pipeline pursues
that standard by construction. Every item
traces to an atomic, ID-tagged knowledge item that records at least two
documented misconceptions at extraction time, and distractors are engineered
from those misconceptions rather than invented when the question is written.
Stem plans map difficulty to coverage. Specifically, easy stems test one knowledge item,
medium stems combine two to three, and hard stems synthesize three to
five. We fix the difficulty distribution up front at 30\% easy, 50\%
medium, 20\% hard, consistent with evidence that explicit difficulty
control steers LLM-generated item pools away from trivially easy
questions~\cite{yao2024mcqg}. Misconception-keyed distractors target exactly the
property that simulation-based item analysis measures as distractor
efficiency~\cite{nguyen2025qgsms}. The pipeline produced a 530-question
assessment bank tagged to a 10-domain, 53-skill blueprint, three
100-question mock exam forms comprising 300 unique items, 96 chapter
quizzes, and ten 70-question simulated practice tests. Quizzes supply
low-stakes retrieval practice while blueprint-matched tests rehearse
certification conditions, and both trace to the same misconception library.

\section{Results}

We report three validation signals, ordered from the most direct evidence:
certification outcomes, capability outcomes, and
independent accreditation. The signals are independent of one another, each
is externally checkable, and none is self-graded which is a property prior
AI-accelerated instructional pipelines have generally
lacked~\cite{yao2025instructional}. Because the deployment is early while involves emerging topics, we
state each sample size plainly and draw no claim beyond what the observed
numbers support.

\subsection{Certification Outcomes}
Three learners prepared for the NVIDIA Certified Professional in Agentic AI
(NCP-AAI) exam using only the framework's knowledge base. All three passed,
a 100\% pass rate to date with $n=3$. The exam is administered and scored by
Nvidia's appointed certification vendor. We control neither its content nor its grading. We
report these outcomes as observed and do not extrapolate the pass rate to
larger cohorts. Three passes establish that studying the knowledge base
alone can carry a learner through a vendor-scored professional
certification. They do not establish the rate at which it will do so.
Fourteen additional learners are progressing toward the exam within the
complete program, and we will report their outcomes as they accrue. Notably, the Nvidia Agentic AI certification (NCP-AAI) is a new certification with currently limited education resources. Most learners are not even aware that this certification exists.

\subsection{Capability Outcomes}
The knowledge base also served as the direct input to a systematic risk analysis of a baseline multi-agent AI system~\cite{nguyen2026mas}. In that analysis, threat-modeling agents worked through the roughly 3,000-page knowledge base chapter by chapter, and produced 1,267 risk items across 81 categories and 14 domains. This signal tests a property that certification exams do not: whether the knowledge base is complete and
well-structured enough to support downstream expert-level analysis, rather
than exam preparation alone. The multi-agent system risk item dataset passed surface validation when it was presented in front of around 500 US federal employees and is being peered reviewed in a historically awarded A* journal.

\subsection{Third-Party Accreditation}
Finally, NASBA reviewed an education program built on the framework's
outputs and approved it for CPE credits. For context,
NASBA National Registry review assesses program design and delivery against
established CPE standards. It is independent of the authors, the
certification vendor, and the adopting agencies. This signal tests the
program as an educational product before a standards body with no stake in
the framework, its vendor alignment, or its adopters.

Taken together, the three signals triangulate the framework from unaffiliated
external directions; each is limited on its own, and their evidential value
lies in their convergence across independent external parties.

\section{Discussion}

These signals matter because of timing. Technical skills carry a half-life
of roughly two and a half years~\cite{daniel2020halflife}, and educational
materials for frontier topics are structurally scarce. When a certification
is new such as the NCP-AAI, preparation corpora, assessment
banks, and trained instructor pools are scarce or absent. Conventional
instructional-design cycles build these artifacts sequentially. For frontier
content the cycle runs slower than the content's decay, and taught programs
add a further stage because instructors must themselves be trained before
they can teach. An AI-accelerated end-to-end pipeline changes this
constraint. Inside the window in which the knowledge is still
current, our framework produced a roughly 3,000-page prerequisite-ordered knowledge base, verified
chapters, 16 tutoring protocols, and a 530-question assessment bank tagged
to a 10-domain, 53-skill blueprint. This is a significant improvement considering prior systems accelerate single stages of this chain. Notably, any unaccelerated stage re-imposes the original bottleneck on the entire
chain.

We also offer an interpretation of why acceleration at this
scale did not cost the framework its grounding. Humans keep the
high-judgment roles in blueprint design, SME review, misconception authoring,
and item rating, while AI absorbs the volume work of drafting,
condensation, and cross-referencing. This division mirrors the leveling
pattern in the studies of Section~I.

The stakes extend beyond individual learners. Beyond the executive
readiness gap noted in Section~I~\cite{mckinseyretraining}, national-scale
exposure far exceeds visible adoption~\cite{chopra2025iceberg}. Certification capacity in frontier
topics is one mechanism by which that latent exposure converts into
workforce readiness rather than displacement. The conversion matters for
distribution as well as output for reasons such as: automation's inequality effects are often
non-monotonic and can rise as technology approaches top human
skill~\cite{benzell2025automation}; economic value increasingly bifurcates
toward verification-grade expertise; and entry-level pressure is already
visible in a reported 16\% relative employment decline for workers aged
22 to 25 in AI-exposed occupations~\cite{catalini2026agi}.

We therefore read the framework not as a convenience for exam candidates but as
infrastructure - a way for institutions and economies to build upskilling
capacity in new technical domains at the speed those domains now change.

\section{Conclusion}

We presented an end-to-end, AI-accelerated, and learning-efficient pipeline that produced a certification-grade training program for a frontier topic, multi-agent AI systems, supported by strong
external signals. All three learners who studied only the knowledge base
passed the NVIDIA Certified Professional in Agentic AI exam. The same
knowledge base seeded a threat-modeling analysis yielding 1,267 risk
items~\cite{nguyen2026mas}; fand NASBA approved an upskilling program built on these
outputs for CPE credits. For organizations facing training gaps in topics too new
to have textbooks, instructors, or item banks, the framework offers a
documented path to produce all three while the knowledge is current.

\bibliographystyle{IEEEtran}
\bibliography{references}

\end{document}